\documentclass[10pt, conference, compsocconf]{IEEEtran}
\IEEEoverridecommandlockouts
\usepackage{cite}
\usepackage{amsmath,amssymb,amsfonts}
\usepackage{algorithmic}
\usepackage{textcomp}
\usepackage{xcolor}

\usepackage{multicol}
\usepackage{times}

\usepackage{soul}
\usepackage[ruled,vlined]{algorithm2e}
\usepackage{url}
\usepackage[hidelinks]{hyperref}
\usepackage[utf8]{inputenc}
\usepackage[small]{caption}
\usepackage{graphicx}
\usepackage{booktabs}
\usepackage{multirow}
\usepackage{xspace}
\usepackage{subfigure}
\newcommand{\bert}{\textsc{Bert}\xspace}

\newcommand{\scibert}{\textsc{SciBert}\xspace}
\newcommand{\biobert}{\textsc{BioBert}\xspace}
\newcommand{\Scierc}{\textsc{S}ci-\textsc{erc}\xspace}
\newcommand{\rdaner}{\textsc{RDANER}\xspace}
\newcommand{\bio}{$\mathcal{BIO}$\xspace}
\newcommand{\cs}{$\mathcal{CS}$\xspace}
\newcommand{\bertcs}{$\bert_{\textsc{CS}}$\xspace}
\newcommand{\bertbio}{$\bert_{\textsc{BIO}}$\xspace}
\newcommand{\rdanercs}{$\rdaner_{\textsc{CS}}$\xspace}
\newcommand{\rdanerbio}{$\rdaner_{\textsc{Bio}}$\xspace}
\newcommand{\bertbase}{$\bert_{\textsc{Base}}$\xspace}

\newcommand{\Vw}{\mathbf{w}}
\newcommand{\Vtheta}{\boldsymbol{\theta}}

\newcommand{\MA}{\mathbf{A}}

\DeclareMathOperator{\softmax}{softmax}

\def\BibTeX{{\rm B\kern-.05em{\sc i\kern-.025em b}\kern-.08em
    T\kern-.1667em\lower.7ex\hbox{E}\kern-.125emX}}

\begin{document}

\title{A Robust and Domain-Adaptive Approach for Low-Resource Named Entity Recognition}

\author{\IEEEauthorblockN{Houjin Yu\IEEEauthorrefmark{1},
Xian-Ling Mao\IEEEauthorrefmark{1},
Zewen Chi\IEEEauthorrefmark{1}, 
Wei Wei\IEEEauthorrefmark{2} and
Heyan Huang\IEEEauthorrefmark{1}}
\IEEEauthorblockA{\IEEEauthorrefmark{1}School of Computer Science,
Beijing Institute of Technology,
Beijing, China}
\IEEEauthorblockA{\IEEEauthorrefmark{2}School of Computer Science,
Huazhong University of Science and Technology,
Hu'bei, China\\
yuhoujin@outlook.com, \{maoxl, czw, hhy63\}@bit.edu.cn, weiw@hust.edu.cn}
}

\maketitle

\begin{abstract}
    Recently, it has attracted much attention to build reliable named entity recognition (NER)
    systems using limited annotated data.
    Nearly all existing works heavily rely on domain-specific resources, such as external lexicons and knowledge bases. 
    However, such domain-specific resources are often not available,
    meanwhile it's difficult and expensive to construct the resources, 
    which has become a key obstacle to wider adoption.
    To tackle the problem, in this work, we propose a novel robust and domain-adaptive approach \rdaner 
    for low-resource NER, which only uses cheap and easily obtainable resources. 
    Extensive experiments on three benchmark datasets demonstrate that
    our approach achieves the best performance when
only using cheap and easily obtainable resources, and delivers competitive results
against state-of-the-art methods which use difficultly obtainable domain-specific resources.
All our code and corpora can be found on \url{https://github.com/houking-can/RDANER}.

\end{abstract}


\section{Introduction}
Named entity recognition (NER) is a fundamental task which is often used as a first step 
in numerous natural language processing (NLP) tasks, including
relation extraction\cite{li2016biocreative,ren2017cotype} and knowledge graph construction\cite{lin2015learning,luan-etal-2018-multi}.
Most existing NER approaches such as neural network-based methods\cite{lample-etal-2016-neural,ma-hovy-2016-end,liu2018empower}, 
often require a large amount of training data (annotated entity spans and types) to achieve satisfactory performance.
It is clearly expensive, and sometimes
impossible, to obtain a large amount of annotated data in a new domain for the NER
task. Under such circumstance, low-resource NER has attracted much deserved attention recently,
which aims to build reliable NER systems using limited annotated data.


Many approaches have been proposed to address low-resource NER.
Early works are mainly based on hand-crafted rules\cite{collobert2011natural,yang2012extracting,luo2015joint}, but they suffer from limited performance in practice. 
More recently, researches on low-resource NER focus on learning information and knowledge from extra domain-specific resources to improve the NER performance.
According to the required resources, they can be divided into two types: 
learning-based methods\cite{luan-etal-2018-multi,zhou-etal-2019-dual,shang-etal-2018-learning} and domain-specific pre-training methods\cite{beltagy-etal-2019-scibert,lee2020biobert}. 
Learning-based methods belong to supervised learning in some sense, such as transfer learning, multi-task learning 
and distantly-supervised learning,
which leverage information and knowledge provided by external lexicons and knowledge bases. 
In fact, it needs extensive amount of experts effort to construct such  resources.
Unlike learning-based methods, domain-specific pre-training methods 
adopt transfer-based pre-training (unsupervised learning) on large amount of in-domain corpora to enable knowledge transfer. 
Domain-specific pre-training methods need less manual effort, but GPU clusters or TPUs (quite expensive) are required to speed up the training process.
Both kinds of methods utilize extra knowledge, either from experts or in-domain corpora,
which have been shown to be effective for low-resource NER.

\begin{table*}[t]    \caption{Summarization of the existing low-resource NER methods.}
    \label{tal:resources}
    \setlength{\tabcolsep}{9pt}

    \centering
    \begin{tabular}{l|c|c}
        \hline
        \multicolumn{1}{c|}{Methods}  & \multicolumn{1}{c|}{Required resources}           & \multicolumn{1}{c}{Limitations}                                                                    \\ \hline
        Transfer learning\cite{zhou-etal-2019-dual,chaudhary-etal-2019-little}                & Parallel corpora,  dictionary & \multirow{3}{*}{\begin{tabular}[c]{@{}c@{}}Difficult to obtain and\\ expensive to construct\end{tabular}} \\
        Multi-task learning\cite{luan-etal-2018-multi,wadden-etal-2019-entity,eberts2019span}           & Annotations for other tasks              &                                                                                                    \\
        Distantly supervised learning\cite{liu-etal-2019-knowledge-augmented,fries2017swellshark,shang-etal-2018-learning} & Knowledge bases, lexicons                &                                                                                                    \\ \hline
        Domain-specific pre-training\cite{beltagy-etal-2019-scibert,lee2020biobert}  & Pre-trained language models             & \begin{tabular}[c]{@{}c@{}}Time-consuming and\\ expensive to pre-train\end{tabular}                \\ \hline
        \end{tabular}
        \end{table*}

Most existing methods for low-resource NER are summarized in Table~\ref{tal:resources}. From the table,
we observe that these methods highly dependent on the availability of domain-specific resources.
However, these resources are often not available, meanwhile it's difficult and expensive to construct them, 
which has become a key obstacle to wider adoption.
For example, it's easy to obtain a general domain knowledge base (like Wikipedia), but we could hardly find a publicly financial knowledge base. 
In fact, it requires large amounts of experts effort and money to build a domain-specific knowledge base.  


To tackle the problem, we propose a novel robust and domain-adaptive approach \rdaner for low-resource NER only using cheap and easily obtainable resources. 
Specifically, the proposed approach consists of two steps: transformer-based language model fine-tuning (LM fine-tuning) and bootstrapping.
Here, LM refers specifically to the transformer-based language model.
Firstly, we fine-tune a general domain pre-trained LM on in-domain corpora to make it
fit on the target domain. Fortunately, it's easy to obtain a general domain pre-trained LM and a large amount of unannotated in-domain corpora.
Then we perform a bootstrapping process, starting from an initial NER model trained on
the small fully-annotated seed data, and then we use it to predict on an unannotated corpus
which is further used to train the model iteratively until convergence.
Our proposed approach alleviates the requirements of difficultly obtainable domain-specific resources, and builds reliable NER systems 
under low-resource conditions, which is a trade-off between effectiveness and efficiency.


To evaluate our proposed approach, we conduct low-resource experiments 
on three benchmark datasets in two challenging domains: computer science and biomedical.
Extensive experiments demonstrate that our proposed approach is not only effective but also efficient.
When only using cheap and easily obtainable resources, our approach outperforms 
baselines with an average improvement of 3.5 F1. Beside, the proposed approach achieves competitive performance against the state-of-the-art
methods which utilize difficultly obtainable domain-specific resources.



    
  

\section{Related Work}

Named entity recognition (NER) has been studied widely for decades. 
Traditionally, NER is concerned with identifying general named entities, such as person, location, and organization names 
in unstructured text. Nowadays, researches have been extended to many specific domains, including biomedical, financial
and academic. Most early NER works are based on hand-crafted rules designed by experts\cite{collobert2011natural,yang2012extracting,luo2015joint}. 
Recently, neural network-based NER models\cite{ma-hovy-2016-end,lample-etal-2016-neural,liu2018empower} have yielded great improvement over the early features-based models, 
meanwhile, requiring little feature engineering and domain knowledge\cite{yang2012extracting,ma-hovy-2016-end}. 
The biggest limitation of such neural models is that they
highly dependent on large amounts of annotated data. As a result, 
the performance of these models degrades dramatically in low-resource settings.
Low-resource NER, which aims to build reliable NER systems, has attracted much attention in recent times.
Researches on low-resource NER
mainly focus on utilizing extra domain-specific resources to improve the performance.
There are mainly two types methods: learning-based methods and domain-specific pre-training methods. 
We introduce the related works of them respectively as follows.

\subsection{Learning-based Methods} 
Learning-based methods for low-resource NER assume some domain-specific resources are available, such as
lexicons, parallel corpora and knowledge bases. 
These methods can be divided into three types: transfer learning (TL), multi-task learning (MTL)
and distantly-supervised learning (DSL).
TL has been extensively used for improving low-resource NER\cite{li2012joint, yang2017transfer}.
Most of them focus on transferring cross-domain knowledge into NER, which
rely on annotation projection methods where annotations in high-resource domains
are projected to the low-resource domains leveraging parallel corpora\cite{chen2010jointly, ni-etal-2017-weakly} and shared representation\cite{zhou-etal-2019-dual,chaudhary-etal-2019-little,cao-etal-2019-low}. 
In fact, many TL methods are designed for general domain NER tasks, 
because it is easier to obtain parallel corpora or bilingual dictionaries from the general domain than 
from a specific domain.

Similarly, MTL utilizes knowledge from extra annotations provided by the dataset, and
adopts jointly training on multiple tasks to help improve NER performance\cite{luan-etal-2018-multi,eberts2019span,wadden-etal-2019-entity}. 
Different from TL, MTL aims at improving the performance of all the tasks instead of low-resource task only. 
The requirements of MTL methods for low-resource NER are manual-labeled annotations for other tasks.

Another trend for better low-resource NER performance is DSL, which has attracted many attentions
to alleviate human efforts. DSL methods use domain-specific dictionaries and knowledge bases to
generate large mount of weakly-annotated data. 
SwellShark\cite{fries2017swellshark} and AutoNER\cite{shang-etal-2018-learning} use dictionary matching 
for named entity span detection. Reference\cite{mathew2019biomedical} combines bootstrapping
and weakly-annotated data augmentation by using a reference set. Their approaches work well only
when domain-specific resources are available.



\subsection{Domain-specific Pre-training Methods}
Transformer-based pre-training
have been shown to be powerful for NLP tasks\cite{devlin2018bert,yang2019xlnet}, including low-resource NER.
But most publicly pre-trained LMs (like GPT, \bert) are trained on general domain corpora, 
they often yields unsatisfactory results in many specific domains. A solution for this problem is domain-specific pre-training,
which trains LMs on in-domain corpora.
\scibert\cite{beltagy-etal-2019-scibert} and \biobert\cite{lee2020biobert}
are two domain-specific \bert variants for scientific text and biomedical text respectively,
showing powerful performance in corresponding domains. 
Different from learning-based methods, pre-training doesn't dependent on
the resources required experts effort to construct. Unfortunately, 
it is quite expensive to pre-train LMs from scratch,
needing GPU clusters or TPUs to speed up the training processes. 

Most existing works heavily rely on difficultly obtainable domain-specific resources, requiring either experts effort or  
high-performance hardware. However, they are often not available or cost lots of money to 
construct them. Without corresponding domain-specific resources, 
it's hard for these methods to be applied in a new domain.
To solve this problem, we propose a novel robust and domain-adaptive approach \rdaner for low-resource
NER, which only uses cheap and easily obtainable researches.

The most related works to ours are semi-supervised methods, which have been explored to further improve the accuracy 
by either augmenting labeled datasets or bootstrapping techniques\cite{he2017unified,chaudhary-etal-2019-little,mathew2019biomedical}.
Reference\cite{chaudhary-etal-2019-little} uses a combination of cross-lingual
transfer learning and active learning for bootstrapping low-resource entity
recognizers. Reference\cite{mathew2019biomedical} combines bootstrapping
and weakly-annotated data augmentation by using an external lexicon to improve NER performance.
Different from \cite{chaudhary-etal-2019-little,mathew2019biomedical}, our proposed approach assumes no
parallel corpora or lexicons in the target domain.

We describe our approach \rdaner in detail as follows.

\section{Approach: \rdaner}

For a specific low-resource NER task, we assume to have 
(1) a small fully-annotated seed dataset $\mathcal{D}_s$ that has every token
tagged by entity type, 
(2) a small in-domain corpus $\mathcal{D}_c$, which is used  to generate weakly-annotated data, 
(3) a general domain pre-trained language model $\mathbf{LM}$ and 
(4) a large-scale in-domain corpus $\mathcal{C}_T$  of the target domain $T$.

\begin{figure}[!t]
    \centering
    \includegraphics[width=0.48\textwidth]{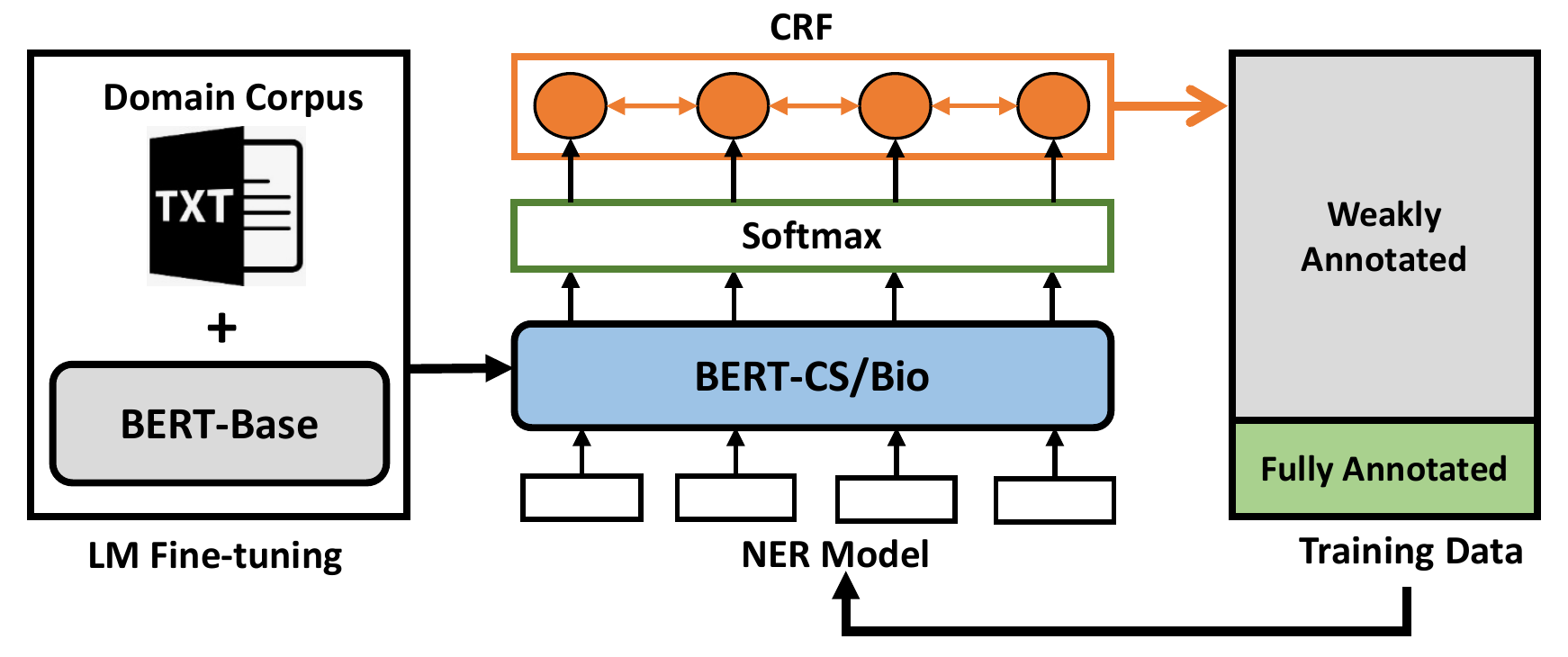}
    \caption{Architecture of \rdaner: LM fine-tuning + Bootstrapping.}
    \label{fig:architecture}
  \end{figure}

As noted in the introduction, \rdaner consists of two processes: LM
fine-tuning and the bootstrapping.
Fig.~\ref{fig:architecture} shows the architecture of the proposed approach. 
We will describe each of them in follow sections.

\subsection{LM Fine-tuning}

Given a general domain pre-trained language model $\mathbf{LM}$ and a domain-specific corpus $\mathcal{C}_T$, 
our goal is to make $\mathbf{LM}$ fit on the target domain $T$.
In this work, $\mathbf{LM}$ is \bert, and we follow the work of pre-training \bert\cite{devlin2018bert}.
Before feeding word sequences into \bert, 
15\% of the words in each sequence are replaced with a {\tt [MASK]} token at random. 
\bert attempts to predict the original value of the masked words, 
based on the context provided by the other non-masked words in the sequence.
The objective is masked language modeling (MLM) cross-entropy\cite{devlin2018bert},
which measures the likelihood of predictions for masked words.
When fine-tuning has completed, we get a \bert variant, $\bert_T$.

\subsection{Bootstrapping}


Bootstrapping is proposed to further improve the accuracy.
First, we train an initial NER model $\mathcal{M}_0$ using the small seed dataset $\mathcal{D}_s$.
We use $\bert_T$ as the word-level encoder, and a linear-chain CRF after a softmax layer.
For an input sequence $ \mathbf{X} = (\mathbf{x}_1, \mathbf{x}_2, ..., \mathbf{x}_n) $ and 
a sequence of tag predictions $ \mathbf{y} = (y_1, y_2, ..., y_n) $, 
$\bert_T$ converts $\mathbf{x}_i$ into a fixed-length vector $\mathbf{w}_i$, and outputs the
probability distributions $\mathbf{h}$ on $\mathbb{R}^K$:
\begin{equation}
    \mathbf{h}= \softmax(\mathbf{w})
    \end{equation}
    where $ \mathbf{w} = (\Vw_1, \Vw_2, ..., \Vw_n) $, 
    $K$ is the number of tags and depends on the the number of classes and on the tagging scheme.
The linear-chain CRF model defines the posterior probability of 
$\mathbf{y}$ given $\mathbf{X}$ to be:
\begin{equation} 
    p(\mathbf{y}|\mathbf{X};\MA) = \frac{1}{Z(\mathbf{X})}\exp \left(\sum_{k=0}^{n} h^{k}(y_{k};\mathbf{X}) + \sum_{k=1}^{n-1} \MA_{y_k,y_{k+1}} \right)
    \end{equation}
    where $Z(\mathbf{X})$ is a normalization factor over all possible tags of $\mathbf{X}$, 
     $h^k(y_k;\mathbf{X})$  indicates the probability of taking the $y_k$ tag at position $k$,
    $y_0$ and $y_{n+1}$ are start and end tags.
    $\MA$ is the transfer matrix, and 
     $\MA_{y_k,y_{k+1}}$ means the probability of 
     a transition from tag states $y_k$ to $y_{k+1}$.
    The most likely tag sequence of $\mathbf{X}$ is represented as follows:
    \begin{equation}
        \mathbf{y}^{*} = \arg\max_{\mathbf{y} \in \mathbb{R}^K} p(\mathbf{y}|\mathbf{X};\MA)
    \end{equation}
     $\MA$ is learnt through the maximum log-likelihood estimation, 
    which maximizes the log-likelihood function $\mathbb{L}$ of 
    training set sequences in seed dataset $\mathcal{D}_s$:
    \begin{equation}
        \mathbb{L}(\mathcal{D}_s, \MA)  =  \sum_{m=1}^{M}\log p(\mathbf{y^{(m)}}|\mathbf{X^{(m)}};\MA) 
    \end{equation}
    where $M$ is the size of the fully-annotated seed dataset $\mathcal{D}_s$.

 Then, we use the initial NER model $\mathcal{M}_0$ to assign labels on $\mathcal{D}_c$, and get a
 weakly-annotated dataset $\mathcal{D}^{*}_{c}$. Moreover,
 we combine $\mathcal{D}_{s}$ with $\mathcal{D}^{*}_{c}$ and get
 an augmented dataset.
 Different from training, we set thresholds $\Vtheta$ to filter tags with low probabilities outputted by the 
 softmax layer. Finally, we iteratively train the NER model with $\mathcal{D}^{*}_{c}$,
 until the model has achieved an acceptable level of accuracy,
or until the maximum number of iterations. 
Algorithm\ref{alg:method} show the overall process of assigning weakly labels,
where {\tt O} stands for none tag.

\begin{algorithm}[h]
    \caption{Weakly labels assignment}
    \label{alg:method}
    \SetAlgoNoLine
        \KwIn{Annotated seed data ($\mathcal{D}_{s}$)}
        \KwIn{Unannotated corpus ($\mathcal{D}_{c}$)}
        \KwOut{NER model ($\mathcal{M}_{K}$)}
        Train NER model $\mathcal{M}_0$ on $\mathcal{D}_{s}$
        \\
        \For{i in 1 \ldots K}{
            $\mathcal{D}^{(i-1)*}_{c} \gets$ Predict using $\mathcal{M}_{i-1}$\\
            $\mathcal{D}^{(i-1)}_{c} \gets$ Relabel $\mathcal{D}^{(i-1)*}_{c}$ \\ \textbf{s.t.}\\
            \If{ $p(\mathbf{y}^{*}_{k}) < \theta$}{
            $ \mathbf{y}^{*}_{k} \gets$ {\tt O}

            }
            Train model $\mathcal{M}_{i}$ on $\mathcal{D}_{s}$ + $\mathcal{D}^{(i-1)}_{c}$
            }
        \Return{$\mathcal{M}_{K}$} 
  \end{algorithm}

\section{Experiments}
We conduct experiments on three benchmark
datasets in two challenging domains to evaluate and compare our proposed approach
with state-of-the art methods.  We further investigate the effectiveness of LM fine-tuning 
and the bootstrapping process respectively.


\subsection{Datasets}

\begin{itemize}
    \item SciERC\cite{luan-etal-2018-multi} annotates entities, 
    their relations, and coreference clusters. 
    Four relations are annotated. 
    \item BC5CDR\cite{li2016biocreative} is from the most 
    recent BioCreative V Chemical and Disease Mention Recognition task. 
    Two relations are annotated.
    \item NCBI-Disease\cite{dougan2014ncbi} focuses on Disease Name Recognition. 
    Only 1 relation is annotated.
\end{itemize}
Table~\ref{tal:datasets} gives the statistics of the datasets used in this work. To be directly comparable with previous works, we
used the official train/dev/test set splits on all datasets.



\begin{table}[t!]
    \centering
    \caption{Datasets overview.}
    \label{tal:datasets}
	\begin{tabular}{c|ccc}
		\toprule
        Seed & \Scierc & BC5CDR  & NCBI-Disease \\
        \hline
        10\% & 189 & 485 & 626 \\
        20\% & 359 & 960 &  1,171\\
        30\% & 540 &  1,348 & 1,737\\
        50\% & 895 &  2,319 & 2,943\\
        100\% & 1,857&  4,611 & 5,825 \\
        \hline
        Domain & CS & Biomedical & Biomedical\\
        \hline
        Entity types & 6 & 2&  1\\
		\bottomrule
    \end{tabular}
\end{table}


\subsection{Cheap and Easily Obtainable Resources}
\begin{itemize}
    \item General Domain Pre-trained LM: We use \bertbase\footnote{\url{https://github.com/google-research/bert}},
        which trained on a general domain corpora including 
        English Wikipedia and BooksCorpus. 
    \item In-domain Corpora: We construct two in-domain corpora, \cs and \bio, to fine-tine \bertbase. 
      The \cs consists 40k papers from AI conference proceedings and 87k papers in AI community from arXiv.
       The \bio consists 200k abstracts which are randomly sampled from PubMed.
\end{itemize}
      
\subsection{Evaluation Metric} Following previous works\cite{zhou-etal-2019-dual,beltagy-etal-2019-scibert,chaudhary-etal-2019-little}, the micro-averaged F1 score is used as
the evaluation metric. 

\subsection{Baselines}
We evaluate our proposed approach \rdaner with \bertbase, which is the strongest baseline in our work.
Then, we compare \rdaner against two domain-specific \bert variants to show the 
effectiveness of LM fine-tuning.
Furthermore, we show the performance of bootstrapping process and
compare \rdaner with state-of-the art learning-based methods which use difficultly obtainable domain-specific resources.
More details are as follows.

\begin{itemize}
    \item \bertbase\cite{devlin2018bert} is a transformer-based language model
        which has been shown to be powerful for general domain NER tasks.

    \item \scibert\cite{beltagy-etal-2019-scibert} is trained on 
        scientific corpora including 18\% papers from the computer science domain
        and 82\% from the broad biomedical domain. 
    \item \biobert\cite{lee2020biobert} is initialized with \bert, then fine-tuned on PubMed abstracts
        and PMC full text articles. 

        \item DyGIE++\cite{wadden-etal-2019-entity} has achieved 
        best performance on NER, especially on \Scierc dataset. However, DyGIE++ requires
        extra annotations including relation, coreference and event labels. 
        \item SpERT\cite{eberts2019span} is an attention model for span-based
        joint entity and relation extraction. It requires relation annotations for training.
    \item SwellShark\cite{fries2017swellshark} is an excellent 
        distantly supervised model in the biomedical domain, which needs no human annotated data.
        However, it requires extra expert effort for designing effective regular expressions.
    \item AutoNER\cite{shang-etal-2018-learning} circumvents the requirements
    of extra human effort, however, it needs large high-quality dictionaries
    to achieve satisfactory performance.

\end{itemize}

\subsection{Implementation Details}
Our model is implemented with AllenNLP\footnote{\url{https://github.com/allenai/allennlp}}.
For all pre-trained \bert variants, we use PyTorch version of them and the 
original code released publicly. All experiments are conducted on a single 
GTX 1080Ti GPU (12GB). 

We fine-tune \bertbase on two large-scale in-domain corpora: \cs and \bio using transformers
library\footnote{\url{https://github.com/huggingface/transformers}}, and get 
two \bert variants \bertcs and \bertbio respectively.
We initialize \bertcs and \bertbio with weights from \bertbase, 
and we use the same vocabulary as \bertbase.
Different from original \bert code, we set a maximum sentence length 
of 120 tokens for \cs corpus and 80 tokens for \bio corpus, 
which are in line with most sentence length in the corresponding corpus.
Both of them are fine-tuned for 1 epoch, and we don't continue train the model 
allowing longer sentence length as \bert does.

To simulate a limited annotated data setting,
we randomly select subsets of training data as seed training datasets with varying data 
ratios at 10\%, 20\%, 30\%, 50\% and 100\%. The remaining training data simulate small in-domain corpora.
Numbers of sentences of each ratio are shown in Table~\ref{tal:datasets}.

The maximum number of iterations is set to 10, and we take the average of last 
5 iterations as the result of each model.
In order to reduce training time, we set different epochs for different seed training datasets, 
epochs decrease as seed increase.  

To investigate the effectiveness of the proposed approach, 
all parameters are fine-tuned on the dev set and we do not perform extensive hyperparameter search. 
For all compared methods, we use the code published in their papers, and follow the same experimental settings.
We add a CRF layer after the softmax layer for all \bert variants.

\section{Results and Discussion}
In this section, we evaluate our proposed approach from the following aspects.
First, we evaluate the proposed approach against \bertbase to investigate the effectiveness of LM fine-tuning process.
Second, we show the performance of the bootstrapping process and investigate the impact of the threshold $\theta$. 
Furthermore, we compare our proposed approach with domain-specific \bert variants and  state-of-the-art 
learning-based methods which use difficultly obtainable domain-specific resources.

\begin{table}
    \caption{NER F1 scores of LM fine-tuning  on \Scierc, BC5CDR and NCBI-Disease with 
    varying training data ratios. 
    }\label{tal:results}
    \centering

    \begin{tabular}{c|c|ccc}
        \toprule
    Dataset      &  Seed  &       \bertbase & LM fine-tuning & \rdaner    \\ \hline
    \multirow{5}{*}{\Scierc}       & 10\%         & 54.07           & 57.39          & \textbf{58.83} \\
                                  & 20\%          & 57.15        & 61.64          & \textbf{62.28} \\
                                  & 30\%          & 60.09        & 63.33          & \textbf{64.61}          \\
                                  & 50\%          & 62.25        & 64.73          & \textbf{65.48}          \\
                                  & 100\%         & 65.24      & 68.46          & \textbf{68.96} \\ \hline
    \multirow{5}{*}{BC5CDR}       & 10\%          & 74.21     & 76.61          & \textbf{78.25}          \\
                                  & 20\%          & 78.43     & 80.51          & \textbf{82.35}          \\
                                  & 30\%          & 79.39      & 81.70          & \textbf{83.55}          \\
                                  & 50\%          & 82.19        & 84.25          & \textbf{85.26}          \\
                                  & 100\%         & 85.61                & 86.87          & \textbf{87.38}          \\ \hline
    \multirow{5}{*}{NCBI-Disease} & 10\%          & 73.03      & 76.09          & \textbf{78.14}          \\
                                  & 20\%          & 79.56       & 80.20          & \textbf{83.46}          \\
                                  & 30\%          & 83.79     & 84.37          & \textbf{85.46}          \\
                                  & 50\%          & 84.88     & 85.70          & \textbf{86.80}          \\
                                  & 100\%         & 86.37      & 87.49          & \textbf{87.89}          \\ 
    \bottomrule
    \end{tabular}
    \end{table}

\subsection{Effectiveness of LM Fine-tuning}

Table~\ref{tal:results} shows the test set evaluation 
results of LM fine-tuning on the three datasets. 
The reported results are the mean across five different runs with different random seeds.
As introduced before, we use \bertcs for \Scierc dataset, \bertbio for BC5CDR and Disease datasets.
We observe that LM fine-tuning consistently outperform \bertbase on all three datasets.
More specifically, LM fine-tuning gains an average improvement of 3.35 F1 
on \Scierc, 2.02 F1 on BC5CDR and 1.24 F1 on NCBI-Disease. It indicates that 
LM fine-tuning on in-domain corpora is effective and domain-adaptive.
We also note that there is performance gap degradation when training data increase for each dataset,
which indicates that LM fine-tuning works better on less training data.
This is due to the text representation has greater impact on NER models
when less training data are provided.





\subsection{Performance of Bootstrapping}

Table~\ref{tal:results} shows the performance of the \rdaner on the three datasets.
\rdaner consists two processes: LM fine-tuning and bootstrapping. 
We observe that the bootstrapping process can always further improve the accuracy of NER models, which indicates
the bootstrapping process is reliable.
Based on LM fine-tuning, the bootstrapping process gains an average improvement of 1.29 F1.
Using only 50\% of training data, \rdaner achieves reasonable performance, which
\bertbase needs 100\% of training data to achieve.

\begin{table*}[t!]
    \caption{Performance of iterative bootstrapping process using 10\% training data. (P: Precision, R: Recall)}
    \label{tal:iterations}
    \centering
    \setlength{\tabcolsep}{9.5pt}
    \small{ 
    \begin{tabular}{@{}cccccccccccc@{}}
    \toprule
    & \multicolumn{3}{c}{\Scierc} && \multicolumn{3}{c}{BC5CDR} && \multicolumn{3}{c}{NCBI-Disease} \\
    \cmidrule{2-4} \cmidrule{6-8}  \cmidrule{10-12}  
      Iter & P & R & F1 && P & R & F1  && P & R & F1\\ 
      \midrule
     0 & 53.84 & 61.44 & 57.39 && 77.42 & 75.82 & 76.61 && 75.62 & 76.56 & 76.09 \\ 
     \hline
     1 & 55.26 & 61.84 & 58.36 && 76.87 & 77.40 & 77.13 && 76.72 & 77.56 & 77.14 \\ 
     2 & 55.34 & 61.62 & 58.31 && 77.13 & 78.65 & 77.88 && 77.23 & 77.71 & 77.47 \\ 
     3 & 55.00 & 62.76 & 58.63 && 78.65 & 76.88 & 77.75 && 76.43 & 77.71 & 77.07 \\ 
     4 & 55.18 & 62.10 & 58.43 && 78.26 & 77.28 & 77.76 && 75.22 & 80.31 & 77.68 \\ 
     5 & 55.14 & 62.22 & 58.47 && 78.41 & 77.14 & 77.77 && 77.04 & 77.60 & 77.32 \\ 
     6 & 55.69 & 62.64 & 58.96 && 77.14 & 77.77 & 77.45 && 78.12 & 77.25 & 77.68 \\ 
     7 & 55.30 & 63.00 & 58.90 && 77.40 & 77.13 & 77.26 && 78.59 & 76.88 & 77.73 \\ 
     8 & 55.23 & 62.16 & 58.49 && 77.77 & 78.26 & 78.01 && 78.16 & 77.19 & 77.67 \\ 
     9 & 56.24 & 62.28 & 59.11 && 76.88 & 77.75 & 77.31 && 78.06 & 77.81 & 77.93 \\ 
     10 & 55.69 & 63.18 & \textbf{59.20} && 77.75 & 78.41 & \textbf{78.08} && 80.33 & 76.56 & \textbf{78.40} \\
    \bottomrule
    \end{tabular}
    }
\end{table*}
In order to get a further understanding of the bootstrapping process,
the iteratively training process with 10\% of the training
data is shown in Table~\ref{tal:iterations}. 
Iteration 0 is the initial model trained on seed dataset, and we use the model to predict labels for unknown tokens repeatedly,
which yields a jump in performance in the first
iteration (Iter 1), since the predicted labels are informative.  
We observe that the bootstrapping process gains an average improvement of 1.86 F1 on initial models across the three datasets. 
The improvement achieved on \Scierc and BC5CDR is mainly due to the gain in recall.
On the contrary, the improvement achieved on NCBI-Disease is mainly due to the gain in precision.
Because there is only 1 type of entity needed to be recognized on NCBI-Disease dataset,
NER on NCBI-Disease is more likely to achieve a high precision.
\begin{figure*}[tb]
    \centering
    \subfigure[\Scierc]{ \includegraphics[width=0.32\textwidth]{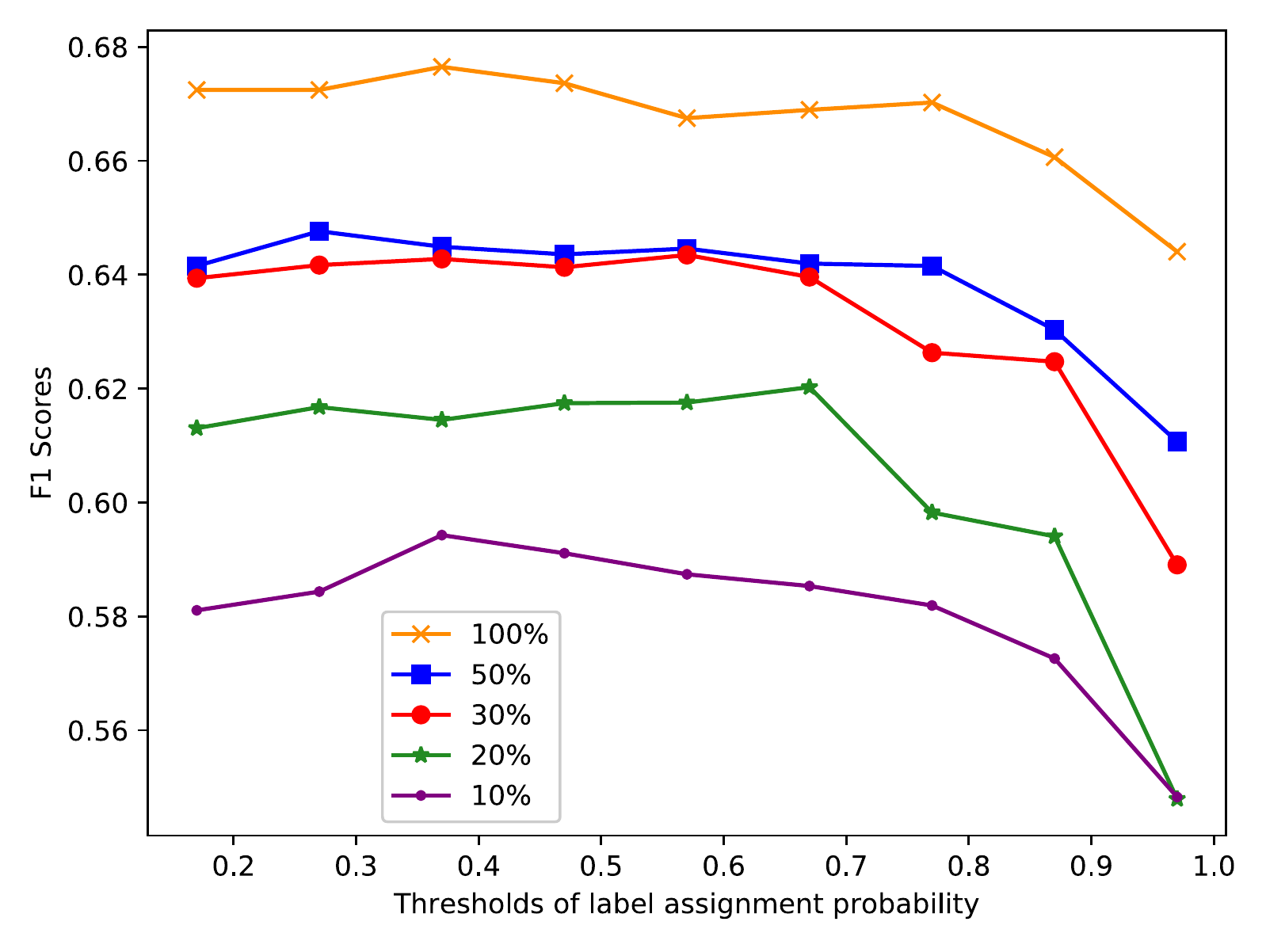}}
    \subfigure[BC5CDR]{ \includegraphics[width=0.32\textwidth]{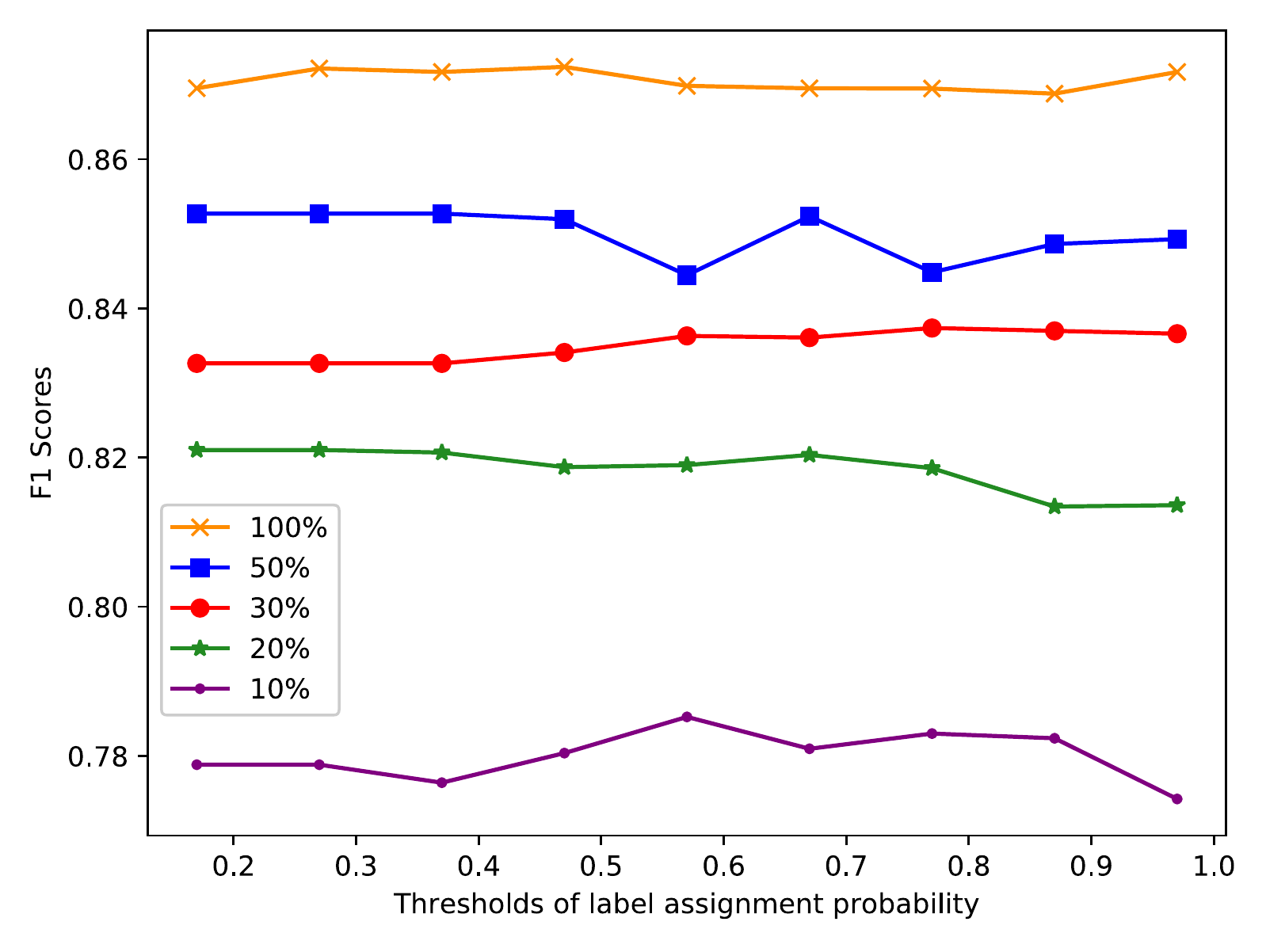}}
    \subfigure[NCBI-Disease]{ \includegraphics[width=0.32\textwidth]{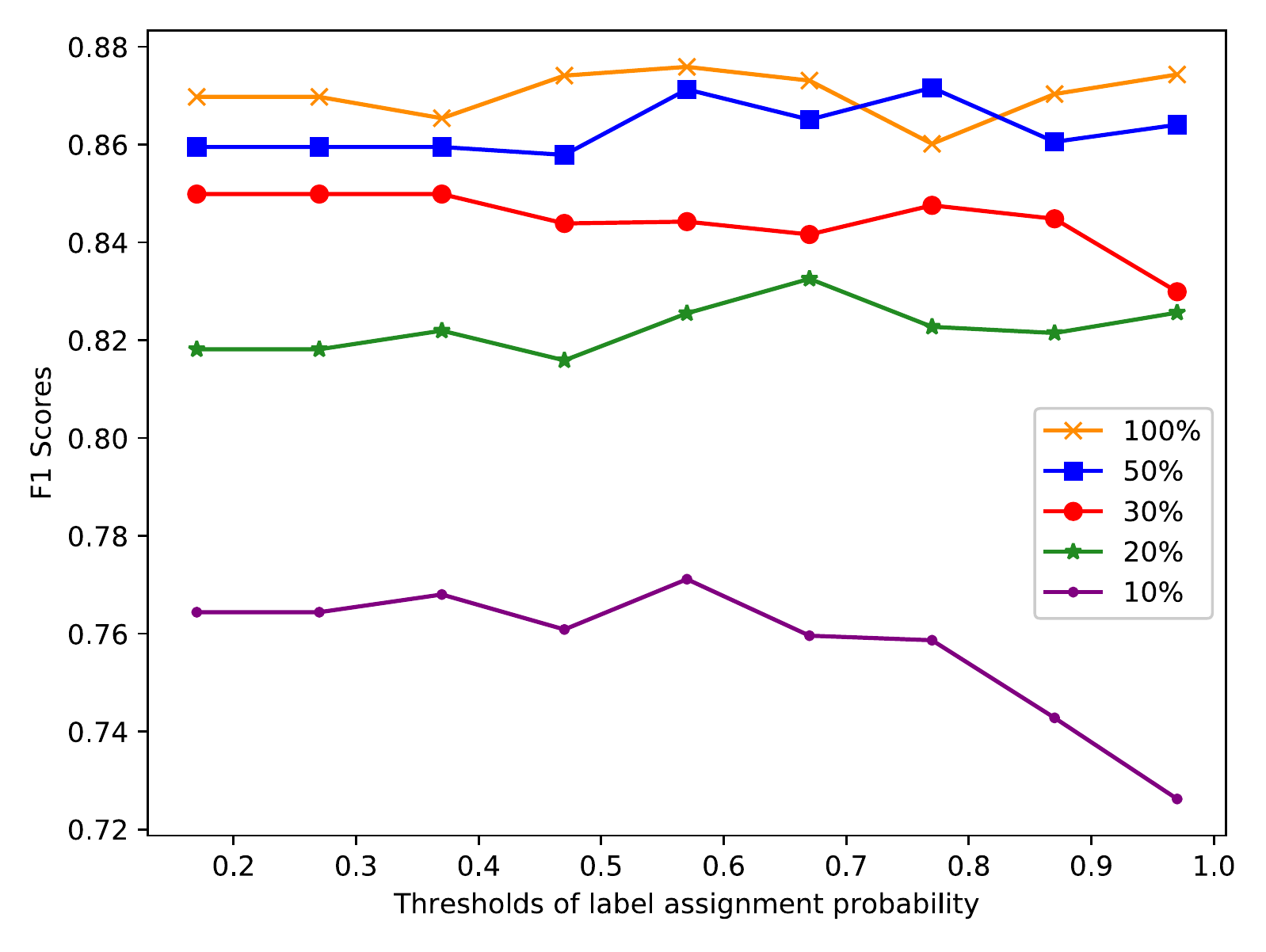}}

    \caption{F1 scores curves on development set vs. thresholds of label assignment
    probability.}
    \label{fig:thresholds}
\end{figure*}

\textbf{Impact of $\Vtheta$:}
Fig.~\ref{fig:thresholds} shows the F1 scores curves on development sets with different thresholds
of label assignment probability. 
This experiment is to find the optimal thresholds
for different ratios of training data.
Theoretically, increasing the threshold does result in a lower number of false positives, leading to 
higher precision. Instead, lower thresholds lead to higher recalls. 
For \Scierc, slowly increasing the threshold results in a slightly more balanced F1 measure.
But the F1 scores start decreasing when precision is higher than a threshold.
We observe that the less training data, the smaller the optimal thresholds, indicating recall 
has a greater impact on F1 under lower resources. Different from \Scierc, curves on BC5CDR and NCBI-Disease are more stable, 
increasing the threshold does little impact on F1 scores except using 10\% of the training data. 
This demonstrates that entity annotated by models have a very high accuracy.
We attribute this to the fact that BC5CDR and NCBI-Disease have fewer entity types,
resulting in high accuracy of classification.

\subsection{Comparison with Domain-specific Pre-training Methods}

\begin{table}[t!]
    \caption{NER F1 scores of domain-specific \bert variants on \Scierc, BC5CDR and NCBI-Disease with 
    varying training data ratios.}\label{tal:domain-specific}
    \centering

    \begin{tabular}{c|c|ccc}
        \toprule
    Dataset  &  Seed   &  \scibert  &  \biobert   &     \rdaner   \\ \hline
                          
    \multirow{5}{*}{\Scierc}       & 10\%             & 57.59          & 57.68          & \textbf{58.83}          \\
                                  & 20\%             & 62.05          & 61.84          & \textbf{62.28}         \\
                                  & 30\%              & \textbf{65.63} & 63.35          & 64.61                  \\
                                  & 50\%                & \textbf{66.95} & 64.66          & 65.48               \\
                                  & 100\%            &  68.55          & 67.89          & \textbf{68.96}         \\ \hline
    \multirow{5}{*}{BC5CDR}       & 10\%               & 81.92          & \textbf{83.56} & 78.25                 \\
                                  & 20\%             & 84.85          & \textbf{85.57} & 82.35                 \\
                                  & 30\%              & 86.14          & \textbf{86.87} & 83.55                  \\
                                  & 50\%               & 87.50          & \textbf{87.94} &      85.26            \\
                                  & 100\%             & \textbf{90.01} & 89.11          &        87.38        \\ \hline
    \multirow{5}{*}{NCBI-Disease} & 10\%                 & 80.82          & \textbf{81.08} &       78.14             \\
                                  & 20\%               & 83.69          & \textbf{85.82} &       83.46         \\
                                  & 30\%                 & 86.35          & \textbf{87.02} &     85.46           \\
                                  & 50\%                 & 87.37          & \textbf{88.53} &   86.80       \\
                                  & 100\%                & 88.57          & \textbf{89.36} &   87.89            \\ 
    \bottomrule
    \end{tabular}
    \end{table}

    \begin{table}[t!]
        \caption{Resources required by domain-specific \bert variants and \rdaner. 
        \rdanercs use \bertcs and \rdanerbio use \bertbio as the text encoder.
        (B: billion, M: million; d: day, h: hour)}
        \label{tal:pre-trained-models}
        \centering
        \setlength{\tabcolsep}{5.5pt}
        \begin{tabular}{c|cc|cc}
            \toprule
    
            Variants & \scibert & \biobert  & \rdanercs & \rdanerbio  \\
            \hline
            Tokens & 3.2B & 18.0B & 17.2M & 37.7M \\
            GPU Time  & 42d & 254d & \textbf{2.5h}  & \textbf{3.5h}\\
    
            \bottomrule
        \end{tabular}
    \end{table}

Table~\ref{tal:domain-specific} 
shows the performance of domain-specific \bert variants on the three datasets.
We observe that \scibert and \biobert perform well on their corresponding domains. \scibert achieves
satisfactory F1 scores on \Scierc (computer science), and 
\biobert performs best on BC5CDR and NCBI-Disease (biomedical). 
Unfortunately, it is very expensive to obtain such domain-specific \bert variants 
because the training processes are computationally expensive, which require high-performance hardware. 
As reported in previous works\cite{beltagy-etal-2019-scibert,lee2020biobert},
it takes 1 week to train \scibert from scratch on a single TPU v3 with 8 cores, 
and 23 days to fine-tune \biobert on eight NVIDIA V100 (32GB) GPUs.
Resources required by domain-specific \bert variants and \rdaner are summarized
in Table~\ref{tal:pre-trained-models}. 
To facilitate comparison, GPU time is converted roughly to the time to train on a single GTX 1080Ti GPU (12GB).
We note that it is very time-consuming to train a domain-specific \bert variant from scratch on a single GPU, 
about 4 months or more. However, \rdaner costs only several hours and achieves satisfactory performance.   
It shows that our proposed approach obtains a trade-off between effectiveness and efficiency.
Compared with domain-specific \bert variants, there is only
a performance drop of 1.87 F1 scores on average for \rdaner. 
That is totally acceptable as it is much cheaper and more efficient.

Surprisingly, \rdaner outperforms \scibert on \Scierc when using 10\%, 20\%, 100\% of the training data. 
Note that \scibert is the best \bert variant in computer science domain currently. 
We attribute this to the fact that \Scierc is a small dataset with only 1,857 annotated sentences.
Unfortunately, bootstrapping is prone to over-fitting on small dataset.
What's more, NER on \Scierc is very challenging because it consists 6 entity types 
and the definitions of entities are ambiguous, such as \textit{Other Scientific Term} and \textit{Method}. 
Unlike \Scierc, BC5CDR and NCBI-Disease are two larger datasets with no more than 2 entity types, thus
NER on the two biomedical datasets is much easier.
\biobert, as the best \bert variant in biomedical domain currently, almost beats all other \bert variants. 
Although there is still a significant gap between \rdaner and \biobert, \rdaner obtain satisfactory
F1 scores costing less money and time. 

Furthermore, we observe that \scibert and \biobert perform similarly on the three datasets. 
The reason is that \scibert is trained on a corpus 82\% from biomedical domain and 18\% from computer science.
It indicates that introducing large-scale in-domain unannotated text for pre-training and fine-tuning
can significantly improve performance. 
In some sense, \biobert is an enhanced version of \bertbio , and the difference between them
is that \biobert uses a much larger corpus than \bertbio (18B tokens vs. 37.7M tokens).
LM fine-tuning gains more improvement with larger in-domain corpora, and we can 
choose the size of in-domain corpora for LM fine-tuning on needs.

\subsection{Comparison with State-of-the-art Learning-based Methods}
\begin{table}[t]
    \caption{NER F1 scores of state-of-the-art learning-based methods using full training data.
    AutoNER and SwellShark require domain-specific lexicons that are unavailable for \Scierc.}
    \label{tal:sota}
    \centering
	\begin{tabular}{lccc}
		\toprule
        Models & \Scierc & BC5CDR  & NCBI-Disease \\
        \hline
        AutoNER\cite{shang-etal-2018-learning} & - & 84.80 & 75.52 \\ 
        SwellShark\cite{fries2017swellshark} & - & 84.23 & 80.80 \\ 
        \hline
        SpERT\cite{eberts2019span} & 67.62 &86.65  & 86.12 \\ 
        DyGIE++\cite{wadden-etal-2019-entity} & \textbf{69.80} & 85.44 & 84.11 \\ 
        \hline
        \rdaner & 68.96 &  \textbf{87.38} &  \textbf{87.71} \\ 
		\bottomrule
    \end{tabular}
\end{table}
In this section, we exploit all training data and use the perfect thresholds of label assignment probability to 
improve the performance of \rdaner.
The averaged results over 5 repetitive runs are summarized in Table~\ref{tal:sota}.
We observe that \rdaner performs reasonably well when compared to various state-of-the-art methods
that use difficultly obtainable domain-specific resources, including distantly supervise learning-base (DSL) methods and
multi-task learning-based (MTL) methods. Because we can't find any parallel 
resources in computer science and biomedical domains, we don't compare
\rdaner with transfer learning methods. 

AutoNER and SwellShark are two DSL methods, and we observe that \rdaner consistently outperforms them.
We can't apply AutoNER and SwellShark on \Scierc dataset, because the domain-specific lexicons are unavailable.  
This also shows DSL methods are highly dependent on the availability of domain-specific resources. 
Although AutoNER and SwellShark claim that they don't use any human annotated data, they actually leverage the information of 
large domain-specific lexicons. For example, AutoNER uses a lexicon contains 322,882 chemical 
and disease entity surfaces, and SwellShark uses ontologies for generating weakly-annotated data. Such resources are often
not available, leading DSL methods less adaptable.
Interesting, we notice that our approach achieves close results to AutoNER and SwellShark using
20\% of the training data (960 sentences) of BC5CDR, and 10\% of the training data (626 sentence) of NCBI-Disease.
Since it is much more challenging to construct big domain-specific lexicons, 
we suggest using \rdaner to build reliable NER systems when domain-specific lexicons are unavailable.

DyGIE++ and SpERT are two latest state-of-the-art MTL methods,
and both of them are built on top of \bert encodings and utilize extra annotations of other tasks. 
DyGIE++ achieves best F1 score on \Scierc, indicating 
that the extra annotations for other tasks, such as relation, event and coreference labels
are helpful to improve performance. 
However, DyGIE++ perform mediocrely on BC5CDR and 
NCBI-Disease dataset, due to the lack of extra annotations of the two datasets.
We note that SpERT performs better than DyGIE++ on BC5CDR and NCBI-Disease.
This is partly due to SpERT only uses relation labels, lacking of extra annotations has
less impact on SpERT than DyGIE++. 
Surprisingly, without extra annotations, \rdaner achieves the second best F1 score on \Scierc.
Besides, \rdaner outperforms DyGIE++ and SpERT on BC5CDR and NCBI datasets.
It demonstrates our proposed approach is not only effective but also domain-adaptive.

\section{Conclusions}
In this paper, we propose a novel robust and domain-adaptive approach \rdaner for low-resource
NER only using cheap and easily obtainable resources. 
We conduct low-resource experiments in two challenging domains and find that:
1) \rdaner is effective and efficient for low-resource NER, 
and it achieves competitive performance against the state-of-the-art methods which utilize
difficultly obtainable domain-specific resources.
2) Beside, \rdaner is domain-adaptive,  which can be easily applied to a new domain.

\section*{Acknowledgments}
The work is supported by National Key R\&D Plan (No. 2018YFB1005100), 
NSFC (No. 61772076, 61751201 and 61602197), NSFB (No. Z181100008918002) 
and the funds of Beijing Advanced Innovation Center for Language Resources (No. TYZ19005).
Xian-Ling Mao is the corresponding author.

\bibliographystyle{IEEEtran}
\bibliography{ickg}

\end{document}